\documentclass{article}
\usepackage{arxiv}
\usepackage{times}
\usepackage{url}            
\usepackage{amsmath}
\usepackage{amssymb}
\usepackage{amsfonts}       
\usepackage{graphicx}
\usepackage{multicol}
\usepackage{tikz}
\usepackage{soul}
\usepackage[hidelinks]{hyperref}
\usepackage{booktabs}
\usepackage{subcaption}

\usepackage{xfrac}
\usepackage{xcolor}

\newcommand{\ie}{\textit{i}.\textit{e}. }
\newcommand{\eg}{\textit{e}.\textit{g}. }

\title{Neural network compression via learnable wavelet transforms}
\author{Moritz Wolter \\
 	    University of Bonn\\
        Fraunhofer Center for Machine Learning and SCAI \\
        {\tt\small wolter@cs.uni-bonn.de}\\
	    \And Shaohui Lin \\
	    National University of Singapore\\
		{\tt\small linsh@comp.nus.edu.sg}
		\And  Angela Yao \\
	    National University of Singapore\\
		{\tt\small ayao@comp.nus.edu.sg}
}

\begin{document}
\maketitle

\begin{abstract}
Wavelets are well known for data compression, yet have rarely been applied to the compression of neural networks. This paper shows how the fast wavelet transform can be used to compress linear layers in neural networks. Linear layers still occupy a significant portion of the parameters in recurrent neural networks (RNNs).  Through our method, we can learn both the wavelet bases and corresponding coefficients to efficiently represent the linear layers of RNNs. Our wavelet compressed RNNs have significantly fewer parameters yet still perform competitively with the state-of-the-art on synthetic and real-world RNN benchmarks \footnote{Source code is available at \url{https://github.com/v0lta/Wavelet-network-compression}}. Wavelet optimization adds basis flexibility, without large numbers of extra weights.
\end{abstract}

\section{Introduction}
Deep neural networks are routinely used in many artificial intelligence applications of computer vision, natural language processing, speech recognition, etc.
However, the success of deep networks has often been accompanied by significant increases in network size and depth.
Network compression aims to 
reduce the computational footprint of deep networks so that they can be applied on embedded, mobile, and low-range hardware.  
%
Compression methods range from quantization~\cite{courbariaux2015binaryconnect,rastegari2016xnor,han2016quantization}, pruning~\cite{han2015learning,lin2018accelerating}, to (low-rank) decomposition of network weights~\cite{denton2014exploiting,lin2016towards}.  

Early methods \cite{denton2014exploiting,han2015learning} separated compression from the learning; compression is performed after network training and then followed by fine-tuning.  Such a multi-stage procedure is both complicated and may degrade performance. Ideally, one should integrate compression into the network structure itself; this has the dual benefit of learning less parameters and also ensures that the compression can be accounted for during learning. 
The more direct form of integrated compression and learning has been adopted in recent approaches~\cite{novikov2015tensorizing,yang2015deep}, typically by enforcing a fixed structure on the weight matrices.  Specifically, the structure must lend itself to some form of efficient projection or decomposition in order to have a compression effect,~\eg ~via circulant projections or tensor train decompositions. Maintaining such a structure throughout learning, however, can be challenging, especially using only the first-order gradient descent algorithms favoured in deep learning. 
Typically, constrained optimization requires managing active and inactive constraints, and or evaluating the Karush-Kuhn-Tucker conditions, all of which can be very expensive. 
We therefore must enforce weight structure differently.

One way to simplify the learning is to fix the projection bases \emph{a priori} \eg~to sinusoids, as per the Fourier transform, or rectangular functions, as per the Walsh-Hadamard transform (WHT) and its derivative the Fastfood transform~\cite{ailon2009fast}. The latter has been used to compress the linear layers of Convolutional Neural Networks (CNNs)~\cite{le2013fastfood,yang2015deep}.  As it relies on non-local basis functions, the Fastfood transform has a complexity of O($n\log n$) for projecting a signal of length $n$. More importantly, however, using fixed bases limits the network flexibility and generalization power. Since the choice of basis functions determines the level of sparsity when representing data, the flexibility to choose a (more compact) basis could bring significant compression gains. We advocate the use of wavelets as an alternative for representing the weight matrices of linear layers.  Using wavelets offers us two key advantages. Firstly, we can apply the fast wavelet transform (FWT), which has only a complexity of $O(n)$ for projection and comes with a large selection of possible basis functions. Secondly, we can build upon the product filter approach for wavelet design~\cite{strang1996wavelets} to directly integrate the learning of wavelet bases as a part of training CNNs or RNNs. Learning the bases gives us added flexibility in representing their weight matrices.

Motivated by these advantages, we propose a new linear layer which directly integrates the FWT into its formulation so layer weights can be represented as sparse wavelet coefficients.
Furthermore, rather than limit ourselves to predefined wavelets as basis functions, we learn the bases directly as a part of network training.  Specifically, we adopt the product filter approach to wavelet design. We translate the two hard constraints posed by this approach into soft objectives, which serve as novel wavelet loss terms. By combining these terms with standard learning objectives, we can successfully learn linear layers by using only the wavelet transform, its inverse, diagonal matrices and a permutation matrix.  As the re-parametrisation is differentiable, it can be trained end-to-end using standard gradient descent.
Our linear layer is general; as we later show in the experiments, it can be applied in both CNNs and RNNs.  We focus primarily on gated recurrent units (GRUs), as they typically contain large and dense weight matrices for computing the cell's state and gate values. Our wavelet-RNNs are compressed by design and have significantly fewer parameters than standard RNNs, yet still remain competitive in performance. Our main contributions can be summarized as follows:
\begin{itemize}
    \item We propose a novel method of \emph{learning FWT basis functions} using modern deep learning frameworks by incorporating the constraints of the product filter approach as soft objectives. By learning local basis functions, we are able to reduce the computational cost of the transform to $O(n)$ compared to existing $O(n \log n)$ methods that use fixed non-local basis functions.
    \item Based on this method, we propose an efficient linear layer which is compressed by design through its wavelet-based representation. This linear layer can be used flexibly in both CNNs and RNNs and allow for large feature or state sizes without the accompanying parameter blow-up of dense linear layers.
    \item Extensive experiments explore the effect of efficient wavelet-based linear layers on the various parts of the GRU cell machinery. Our approach demonstrates comparable compression performance compared to state-of-the-art model compression methods.
\end{itemize}

\section{Related Work}

\subsection{Structured Efficient Linear Transforms}
Our proposed approach can be considered a structured efficient linear transform, which replaces unstructured dense matrices with efficiently structured ones.  There are several types of structures, derived from fast random projections~\cite{ailon2009fast}, circulant projections \cite{cheng2015exploration,araujo2018training}, tensor train (TT) decompositions~\cite{novikov2015tensorizing,tjandra2017compressing,yang2017tensor}, low-rank decompositions \cite{denil2013predicting,denton2014exploiting,jaderberg2014speeding}

One of the main difficulties in using structured representation is maintaining the structure throughout learning. One line of work avoids this by simply doing away with the constraints during the learning phase. For example, low-rank decompositions~\cite{denil2013predicting,denton2014exploiting,jaderberg2014speeding} split the learned dense weights into two low-rank orthogonal factors.
The low-rank constraint then gradually disappears during the fine-tuning phase.  The resulting representation is uncontrolled, and must trade off between the efficiency of the low rank and effectively satisfying the fine-tuning objective.  In contrast, our proposed soft constraints 
can be applied jointly with the learning objective, and as such, not only does not require fine-tuning, but can ensure structure throughout the entire learning process.  

Within the group of structured efficient linear transforms, the one most similar to the FWT that we are using is the Fastfood transform~\cite{le2013fastfood}. 
The fastfood transform is applied to reparameterize linear layers as a combination of 5 types of matrices: three random diagonal matrices, a random permutation matrix and a Walsh-Hadamard matrix.
However, these three diagonal matrices are fixed after random initialization, resulting in a non-adaptive transform. Its fixed nature limits the representation power. As a remedy,~\cite{yang2015deep} proposed an adaptive version in which the three diagonal matrices are optimized through standard backpropagation. 
Nevertheless, the approach still uses a fixed Walsh-Hadamard basis 
which may limit the generalization and flexibility of the linear layer.
In contrast to the adaptive Fastfood transform, our method is more general and reduces computational complexity. 

\subsection{Compressing Recurrent Neural Networks}

Compressing recurrent cells can be highly challenging; the recurrent connection forces the unit to be shared across all time steps in a sequence so minor changes in the unit can have a dramatic change in the output. To compress RNNs, previous approaches have explored pruning~\cite{wen2017learning,narang2017exploring,wang2019simultaneously}, quantization~\cite{wang2017accelerating} and structured linear transforms~\cite{tjandra2017compressing,yang2017tensor,pan2019compressing,ye2018learning}. 

The use of structured efficient linear transforms for compressing RNNs has primarily focused on using tensor decompositions, either via tensor train decomposition~\cite{tjandra2017compressing,yang2017tensor} or block-term tensor decomposition~\cite{ye2018learning}. The tensor decomposition replaces linear layers with tensors with a lower number of weights and operations than the original matrix.  Tensor train 
decomposition can compress the input-to-hidden matrix in RNNs, but requires 
the restricted setting on the hyperparameters (\eg ranks and the restrained order of core tensors) making the compressed models sensitive to parameter selection \cite{tjandra2017compressing}.
Pan \emph{et al.} ~\cite{pan2019compressing} employ low-rank tensor ring decomposition to alleviate the strict constraints in tensor train decomposition.
However, these methods need to approximate the matrix-vector operation by tensor-by-tensor multiplication, where the dense weights and input vectors are reshaped into higher-order tensors. 
This requires additional reshaping time and generates feature-agnostic representations during training.
In contrast, our method shows more flexibility and efficiency performing on the fast wavelet transform, whose bases satisfy with wavelet design using soft objectives. 

\subsection{Wavelets in Machine learning}
A body of works using wavelets in machine learning exists. A group of publications is exploring how wavelets can be used to process weighted graphs trough wavelet-autoencoders \cite{rustamov2013wavelets} or convolutions with wavelet-constraints \cite{bruna2013spectral}. In deep learning wavelets have been used \eg  ~as input features \cite{chen2015automatic} or as a tool for weight representation. The latter category includes the definition of convolution filters in the wavelet domain~\cite{cotter2018deep}. We define a new wavelet based structured efficient linear transform which replaces large dense layers. Using our new transform greatly reduces network size as we show in the experimental section.

\section{Method}
\subsection{Fast Wavelet Transform}
The wavelet transform projects signals into a multi-resolution spectral domain defined by the wavelet basis of choice.  The wavelets themselves are oscillating basis functions derived from scaling and translating a prototypical mother wavelet function such as the Haar, Mexican hat, Daubechies, etc. We refer the curious reader to the excellent primer~\cite{strang1996wavelets} for a thorough treatment on the topic.  For our purposes, we can consider the wavelet transform as being analogous to the Fourier transform.  Similarly, wavelets are akin to sinusoids, with a key distinction however, that wavelets are localized basis functions, \ie are not infinite.  

Similar to the fast Fourier transform, there exists a fast wavelet transform $\mathcal{W}$ and an inverse $\mathcal{W}^{-1}$ which can be expressed as matrix multiplications \cite{strang1996wavelets}:
\begin{align}
\mathcal{W}(\mathbf{x}) = \mathbf{A}\mathbf{x} = \mathbf{b},\\ \mathcal{W}^{-1}(\mathbf{b}) = \mathbf{S}\mathbf{b} = \mathbf{x}.
\end{align}

\begin{figure}[t]
    \centering
    \resizebox{.6\linewidth}{!}{
\begin{tikzpicture}
  \node[draw, rectangle] at (0,1) (x){$\mathbf{x}$};
  \node[draw, rectangle] at (1.5,0) (hh){$\mathbf{H_1}\; \downarrow 2$ };
  \node[draw, rectangle] at (1.5,2) (hl){$\mathbf{H_0}\; \downarrow 2$};
  
  \node[draw, rectangle] at (3,1) (hhh){$\mathbf{H_1} \; \downarrow 2$};
  \node[draw, rectangle] at (3,3) (hll){$\mathbf{H_0} \; \downarrow 2$};

  \node[draw, rectangle] at (3,0) (bj0){$\mathbf{b}_{j}$};
  \node[draw, rectangle] at (4.5,1) (bj1){$\mathbf{b}_{j-1}$};

  \node[draw, rectangle] at (6.5,1) (fhh){$\mathbf{F_1}\; \uparrow 2 $};
  \node[draw, rectangle] at (6.5,3) (fll){$\mathbf{F_0}\; \uparrow 2$};
  
  \node[draw, rectangle] at (8,0) (fh){$\mathbf{F_1} \uparrow 2\;$};
  \node[draw, rectangle] at (8,2) (fl){$\mathbf{F_0} \uparrow 2\;$};
  \node[draw, rectangle] at (9,1) (xhat){$\hat{\mathbf{x}}$};
  
  \draw[->] (x) |- (hh);
  \draw[->] (x) |- (hl);
  \draw[->] (hl) |- (hhh);
  \draw[->] (hl) |- (hll);

  \draw[->] (fll) -| (fl);
  \draw[->] (fhh) -| (fl);
  \draw[->] (fl) -| (xhat);
  \draw[->] (fh) -| (xhat);

  \draw[->] (hh) -- (bj0);
  \draw[->] (hhh) -- (bj1);

  \draw[->, dotted] (hll) -- (fll);
  \draw[->] (bj1) -- (fhh);
  \draw[->] (bj0) -- (fh);

\end{tikzpicture}
    }
    \caption{Efficient wavelet signal analysis and synthesis following a tree structure. ~\cite{strang1996wavelets}.
    $\mathbf{H}$ denotes analysis filters and $\mathbf{F}$ stands for synthesis filters. 
     Up~($\uparrow$) and down~($\downarrow$)-
    sampling by a factor of two is written as the arrow followed by the factor. Filtering and sampling can be accomplished jointly in deep learning frameworks by using strided convolutions for analysis and strided transposed convolutions for synthesis. In place of the dotted arrow more scale-levels can be included.}
    \label{fig:fwt}
\end{figure}
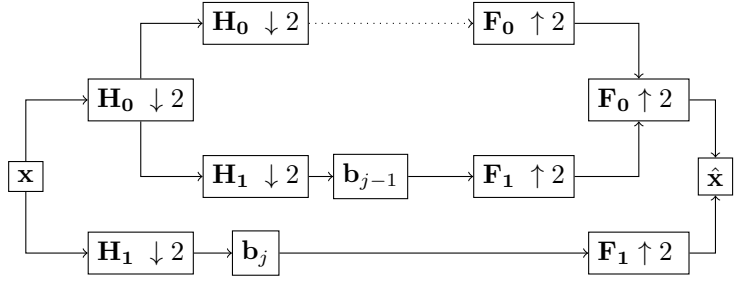

Given a signal $\mathbf{x}$ indexed by $n$, the forward wavelet transform yields coefficients $b_{jk}$ in vector $\mathbf{b}$, which projects $\mathbf{x}$ onto the wavelet basis $\mathbf{A}$. Matrices $\mathbf{A}$ and $\mathbf{S}$ are referred to as analysis (or forward) and synthesis (or backward) matrices respectively.  They are inverses of each other, \ie~$\mathbf{A} =\mathbf{S}^{-1}$, and allow for reconstruction of the signal from the wavelet coefficients. The double indices $jk$ of $b_{jk}$ denote the scale $j$ and time $k$ positions of each coefficient.
Coefficients $b_{jk}$ can be found by recursively convolving $\mathbf{x}$ with the analysis filters $\mathbf{h}_0$ and $\mathbf{h}_1$ with a stride of 2. This process is depicted in Figure~\ref{fig:fwt}. The number of scales ($6$ in our case) is chosen depending on the problem. As a result of the strided convolution, the number of time steps is halved after each scale step.
By working backwards through the scales, one can reconstruct $\hat{\mathbf{x}}$ from $b_{jk}$ through transposed convolutions with the synthesis filters $\mathbf{f}_0$ and $\mathbf{f}_1$.

\subsection{Learning Wavelet Bases}\label{sec:learnable-wavelets}
Typically, wavelets basis functions are selected from  
a library of established wavelets, \eg as the rectangular functions of Haar wavelet, or arbitrarily designed by hand by the practitioner. Based on the product filter approach, designed wavelets must fulfill conditions of perfect reconstruction and alias cancellation~\cite{strang1996wavelets}. 
Given filters $\mathbf{h}$ and $\mathbf{f}$ as well as their z-transformed counterparts $H(z) = \sum_n \mathbf{h}(n) z^{-n}$ and $F(z)$ respectively, the reconstruction condition can be expressed as 
\begin{equation}
H_0(z) F_{0}(z) + H_1(z) F_1(z) = 2z^l,
\label{eq:pr_condition}
\end{equation}
and the anti-aliasing condition as
\begin{equation} \label{eq:ac}
H_0(-z) F_{0}(z) + H_1(-z) F_1(z) = 0.
\end{equation}
For the perfect reconstruction condition in Eq.~\ref{eq:pr_condition}, the center term $z^l$ of the resulting z-transformed expression must be a two; all other coefficients should be zero. $l$ denotes the power of the center. 
To enforce these constraints in a learnable setting, we can design corresponding differentiable loss terms which we call a wavelet loss.
Instead of working in the $z-$space, 
we leverage the equivalence of polynomial multiplication and coefficient convolution ($*$) and reformulate Eq. ~\ref{eq:pr_condition} 
as:
\begin{align} \small \label{eq:lpr}
\mathcal{L}_{pr}(\theta_w) = \sum_{k={0}}^{N} \Big[(\mathbf{h}_0*\mathbf{f}_0)_k + (\mathbf{h}_1*\mathbf{f}_1)_k - \mathbf{0}_{2, k}\Big]^2,
\end{align}
where $\mathbf{0}_2$ is a zero vector with a two at $z^l$ in its center. 
This formulation amounts to a measure of the coefficient-wise squared deviation from the perfect reconstruction condition.
For alias cancellation, we observe that Eq.~\ref{eq:ac} is satisfied if $F_0(z) = H_1(-z)$ and $F_1(z) = -H_0(-z)$ and
formulate our anti-aliasing loss as:
\begin{align} \label{eq:lac}
\mathcal{L}_{ac}(\theta_w) = \sum_{k={0}}^{N} \Big(f_{0,k} - (-1)^k h_{1, k}\Big)^2 + \Big(f_{1, k} + (-1)^k h_{0, k}\Big)^2 .
\end{align}
This formulation leads to the common alternating sign pattern, which we will observe later. We refer to the sum of the two terms in Eqs.~\ref{eq:lpr} and~\ref{eq:lac} as a wavelet loss.  It can be added to standard loss functions in the learning of neural networks.

\subsection{Efficient Wavelet-based Linear Layers}
To use the wavelet-based linear layer, we begin by decomposing the weight matrices $\mathbf{W}$ as follows:
\begin{align} 
\mathbf{W} = \mathbf{D}\mathcal{W}^{-1}\mathbf{G}\mathbf{\Pi}\mathcal{W}\mathbf{B},
\label{eq:wavlet-basis}
\end{align}
where $\mathbf{D}, \mathbf{G}, \mathbf{B}$ are diagonal learnable matrices of size $n \times n$, and $\mathbf{\Pi} \in \{0, 1\}^{n \times n}$ is a random permutation matrix, which stays fixed during training. We use identity matrices as initialization for $\mathbf{D}, \mathbf{G}, \mathbf{B}$. $\mathcal{W}$ and $\mathcal{W}^{-1}$ denote the wavelet transform and it's inverse, which can also be optimized during training. This approach is similar to~\cite{yang2015deep}, which relies on the fast Welsh-Hadamard transform.
$\mathbf{D}, \mathbf{G}, \mathbf{B}$ and $\mathbf{\Pi}$ can be evaluated in $O(n)$~\cite{arjovsky2016unitary}. 
The fast wavelet transform requires only $O(n)$ steps instead of the $O(n \ln n)$ used by the non-local fast Fourier and fast Welsh-Hadamard transforms \cite{strang1996wavelets}. 
This is asymptotically faster than the transforms used in \cite{yang2015deep} and \cite{arjovsky2016unitary}, who work with fixed Welsh-Hadamard and Fourier transforms.
Non square cases where the number of inputs is larger than $n$ 
can be handled by concatenating square representations with tied wavelet weights~\cite{yang2015deep}.

We can replace standard weights in linear layers with the decomposition described above, and learn the matrices in Eq. \ref{eq:wavlet-basis} using the wavelet loss as defined in Eq. \ref{eq:lpr} and \ref{eq:lac} jointly with the standard network objective.  Given the network parameters $\theta$ and all filter coefficients $\theta_w$ we minimize:
\begin{align}
\min(\mathcal{L}(\theta)) = \min[(\mathcal{L}_{o}(\theta)) + \mathcal{L}_{pr}(\theta_w) + \mathcal{L}_{ac}(\theta_w)],
\end{align}
where $\mathcal{L}_{o}(\theta)$ the original loss function (\eg cross-entropy loss) of the uncompressed network, and $\mathcal{L}_{pr}(\theta_w) + \mathcal{L}_{ac}(\theta_w)$ the extra terms for the learnable wavelet basis. 

Our wavelet-based linear layer can be applied to fully-connected layers in CNNs and RNNs.  For example, suppose we are given a gated recurrent unit (GRU) as follows:
\begin{align}
\mathbf{g}_r &= \sigma(\mathbf{W}_r \mathbf{h}_{t-1} + \mathbf{V}_r \mathbf{x}_t + \mathbf{b}_r ), \label{eq:reset} \\
\mathbf{g}_z &= \sigma(\mathbf{W}_z \mathbf{h}_{t-1} + \mathbf{V}_z \mathbf{x}_t + \mathbf{b}_z ), \label{eq:update}\\
\mathbf{z}_t &= \mathbf{W}(\mathbf{g}_{r} \odot \mathbf{h}_{t-1}) + \mathbf{V}\mathbf{x}_t + \mathbf{b} \label{eq:state},\\
\mathbf{h}_t &= \mathbf{g}_z \odot \tanh(\mathbf{z}_t) + (1 - \mathbf{g}_z) \odot \mathbf{h}_{t-1},
\end{align}
where $\sigma(\cdot)$ and  $\text{tanh}(\cdot)$ are sigmoid and tanh activation functions, $\mathbf{z}_t$ is the candidate hidden layer values, $\mathbf{h}_t$ is the hidden layer state at the $t$-th time, and $\mathbf{g}_r, \mathbf{g}_z$ are the reset and update gates, respectively. $\odot$ denotes element-wise multiplication.
$\mathbf{V}$, $\mathbf{V}_r$ and $\mathbf{V}_z$ are the input weights. We can learn efficient gating units by applying the representation from Eq.~\ref{eq:wavlet-basis} to the recurrent weight matrices $\mathbf{W}$, $\mathbf{W}_r$ and $\mathbf{W}_z$. The recurrent weight matrices are of size
$n_h \times n_h$ and are typically the largest matrices in a GRU and learning efficient versions of them can reduce the number of network parameters up to $90\%$. 
%
\begin{figure}[t]
\centering
\begin{subfigure}[t]{0.45\textwidth}
\includegraphics[width=\linewidth]{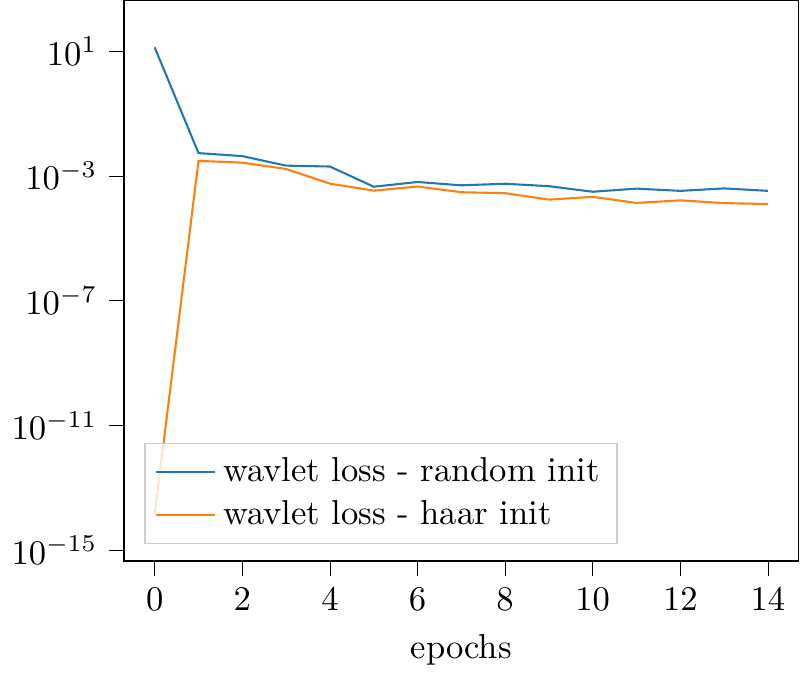}
\caption{Wavelet loss}
\label{fig:random_init}
\end{subfigure}
\begin{subfigure}[t]{0.45\textwidth}
\includegraphics[width=\linewidth]{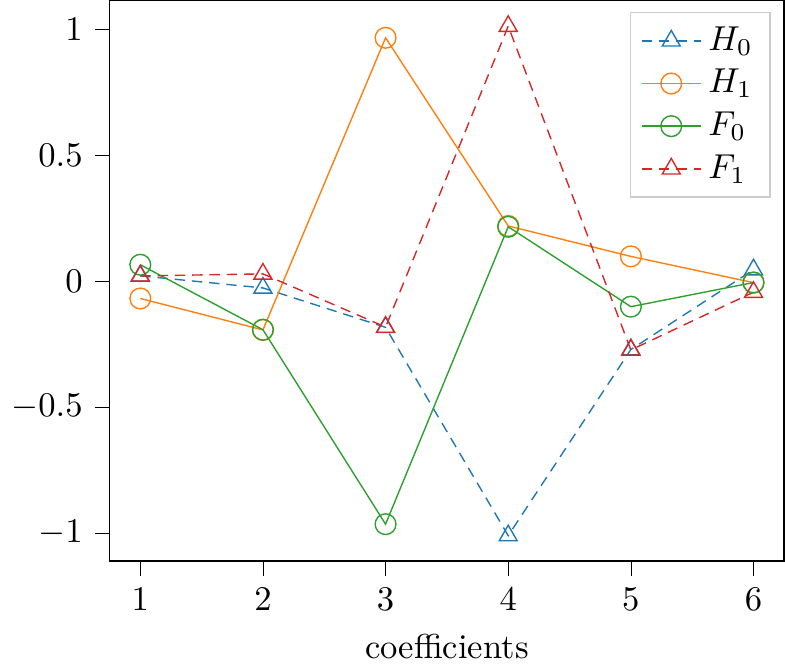}
\caption{Learned wavelet filter coefficients}
\label{fig:learned_filters}
\end{subfigure}
\caption{(a) Wavelet loss sum of a randomly and Haar initialized wavelet array. In both cases, filter values converge to a product filter as indicated by trend of the wavelet loss towards zero . (b) Learned wavelet filter coefficients. Coefficients have been initialized at random. After training, the effects of the alias cancellation constraint are prominently visible.  We must have $F_0(z) = H_1(-z)$ and $F_1(z) = -H_0(-z)$ for alias to cancel itself. Inserting $(-z)$ into the coefficient polynomial leads to a minus sign at odd powers. Additional multiplication with $(-1)$ shifts it to even powers. Alias cancellation therefore imposes an alternating sign pattern. When $F_0$ and $H_1$ share the same sign $F_1$ and $H_0$ do not and vice versa.}
\end{figure}

\section{Experiments}
We evaluate the effectiveness of our linear layer in both CNNs and RNNs. For CNNs, we select LeNet-5 on MNIST digit recognition classification benchmark \cite{lecun1998gradient} as a baseline. For RNNs, we test several GRU models on sequence learning tasks including the copy-memory and adding problem~\cite{hochreiter1997long}, sequential MNIST  and Penn-Treebank (PTB) character modelling~\cite{marcus1993building}.

\begin{table}[t]
\centering
\caption{Experimental results on the MNIST digit-recognition. We work with a LeNet architecture as proposed in previous work. In comparison to the Fastfood approach~\cite{yang2015deep} we obtain comparable performance with slightly fewer parameters. The size of our learnable-wavelet compression layer is set to 800.} 
\begin{tabular}{c c c c}
Net             & Error & Parameters & Reduction \\ \toprule
LeNet-5          & 0.87\%   & 431K  & -  \\
LeNet-Fastfood  & 0.71\%   & 39K & 91\%  \\
LeNet-random    & 1.67\%   & 36K & 92\% \\
LeNet-Wavelet   & 0.74\%   & 36K & 92\%  \\ \bottomrule
\end{tabular}
\label{tab:wavelet-mnist}
\end{table}

\subsection{MNIST-Digit Recognition}\label{seq:cnn_mnist}
We first apply our efficient wavelet layers to the MNIST digit recognition problem~\cite{lecun1998gradient}. MNIST has 60K training and 10K test images with a size of $28\times28$ from 10 classes.
We train using an Adadelta optimizer for 14 epochs using a learning rate of 1. The adaptive Fastfood transform is replaced with our proposed learnable-wavelet compression layer. In this feed-forward experiment, we apply dropout to the learned diagonal matrices $\mathbf{D}, \mathbf{G}, \mathbf{B}$ in Eq. ~\ref{eq:wavlet-basis}.

We start by randomly initializing the wavelet array, consisting of 6 parameters per filter, with values drawn from the uniform distribution $\mathcal{U}_{-1}^1$. In this case, the initial condition is not a valid product filter so the wavelet loss is initially very large, as shown in  Figure~\ref{fig:random_init}. However, as this loss term decreases as the random values in the wavelet array start to approximately satisfy the conditions of Eq. \ref{eq:pr_condition} and Eq. \ref{eq:ac}.  Correspondingly, the accuracy also rises. With the random initialization we achieve a recognition accuracy of $98.33 \%$ with only 36k parameters. We visualize the learned filters in Figure~\ref{fig:learned_filters}. The alias cancellation condition causes an alternating sign pattern. Whenever $F_0$ and $H_1$ have the the same sign $F_1$ and $H_0$ do not and vice versa.
As we observe high recognition rates and small wavelet loss values, we are confident that we have learned valid wavelet basis functions. 

Inspired by the Welsh-Hadamard matrix, we also test with a zero-padded Haar wavelet as a initialization; this should ensure a valid FWT at all times.  In this case, the initial wavelet loss is very small as shown in Figure~\ref{fig:random_init}, as we already start with a valid wavelet that perfectly satisfies conditions~\ref{eq:pr_condition} and ~\ref{eq:ac}.  However, as learning progresses, the condition is no longer satisfied, hence the jump in the loss, before the gradual decrease once again.    
In Table~\ref{tab:wavelet-mnist}, we compare our result to the Fastfood transform \cite{yang2015deep}. Our method achieves a comparable result with a higher parameter reduction rate of 92\% (\emph{vs.} 91\%).

\begin{table}[t]
\centering
\caption{RNN compression results on the adding and memory problems, exploring the impact of our efficient wavelet-based linear layer at various locations in the GRU. On the adding problem all tested variants are functional. Compressing the state and reset equations has virtually no effect on performance. Compressing the update gate leads to a working cell, but cells with a compressed update gate perform significantly worse. Note that on the adding problem, predicting a sum of 1 regardless of the input leads to an mse of 0.167. 
On the copy-memory benchmark, replacing the the state and reset weight matrices with our efficient wavelet version is possible without significant performance losses. A state size of 512 was used for all models. The expected cross entropy for a random guess is 0.12 with n=8.}
\begin{tabular}{ c c c c c c c c c c}
     & &\multicolumn{3}{c}{Adding problem}                               & & \multicolumn{3}{c}{Copy-memory problem}  & \\ \toprule
     &   reduced   & mse-loss     & accuracy      & weights           & &    ce-loss     & accuracy      & weights \\ \midrule
     GRU  &    -           & 4.9e-4  & 99.23\%    & 792K              & &  4.8e-5   & 99.99\% & 808K \\
Wave-GRU & $\mathbf{z_t}$  & 3.0e-4    & 99.45\%  & 531K              & & 2.4e-3  & 99.1\%  & 548K  \\
Wave-GRU & $\mathbf{g_r}$  & 1.1e-4    & 99.96\%  & 531K              & & 3.7e-5  & 99.98\% & 548K  \\
Wave-GRU & $\mathbf{g_z}$  & 4.4e-4    & 97.78\%  & 531K              & & 1.1e-1  & 21.63\% & 548K  \\
Wave-GRU & $\mathbf{g_r}, \mathbf{g_z}$ & 0.9e-4   & 99.85\%  & 270K  & & 3.7e-2 & 73.63\%  & 288K  \\
Wave-GRU & $\mathbf{z_t}$, $\mathbf{g_r}$& 3.0e-4  & 98.56\%  & 270K  & & 2.4e-3   & 99.05\% & 288K \\
Wave-GRU & $\mathbf{z_t}$, $\mathbf{g_z}$& 1.1e-3  & 92.64\%  & 270K  & & 1.2e-1  & 12.67\%  & 288K \\
Wave-GRU & $\mathbf{z_t}$, $\mathbf{g_r}$, $\mathbf{g_z}$  & 1.0e-3 & 91.64\%  & 10K & & 1.2e-1 & 16.84\%  & 27K   \\
Ff-GRU   & $\mathbf{z_t}$, $\mathbf{g_r}$, $\mathbf{g_z}$  & 1.3e-3 & 85.99\%  & 10K & & 1.2e-1 & 16.44\%  & 27K  \\ \bottomrule
\end{tabular}
\label{tab:adding_memory}
\end{table}

To explore the effect of our method on recurrent neural networks, we consider the challenging copy-memory and adding tasks as benchmarks~\cite{hochreiter1997long}.

The copy memory benchmark consists out of a sequence of 10 numbers, $T$ zeros, a marker and 10 more zeros. The tested cell observes the input sequence. It must learn to reproduce the original input sequence after the marker. The numbers in the input sequence are drawn from $\mathcal{U}_0^n$ . Element $n+1$ marks the point where to reproduce the original sequence. We choose to work with n=8 in our experiments and use a cross entropy loss. Accuracy is defined as the percentage of correctly memorized integers.
For the adding problem, $T$ random numbers are drawn from $\mathcal{U}_0^1$ out of which two are marked. The two marks are randomly placed in the first and second half of the sequence. After observing the entire sequence, the network should produce the sum of the two marked elements. A mean squared error loss function is used to measure the difference between the true and the expected sum. We count a sum as correct if $|\hat{y} - y| < 0.05$.
We test on GRU cells with a state size of 512.  $T$ is set to 150 for the adding problem. 
Optimization uses a learning rate of 0.001 and Root Mean Square Propagation (RMSProp).

We first explore the effect of our efficient wavelet layer on the reset $g_r$ and update $g_z$ equations (Eq. \ref{eq:reset}, \ref{eq:update}) as well as the state $z_t$ equation (Eq. \ref{eq:state}). As shown in Table~\ref{tab:adding_memory},
We observed that efficient representations of the state and reset equations has little impact on performance, while significantly reducing the weights. In the combined case, our method has 2.8$\times$ less parameters than dense weight matrices with only 0.91\% accuracy drop in copy memory problem. For the adding problem, our method achieves a factor of 2.9$\times$ reduction with only 0.67\% accuracy drop.  When incorporating this into multiple weight matrices, we find that using the efficient representation is problematic for the update matrix $W_z$ next state computation.  
We found that compressing the update gate has a large impact on the performance, especially the combination with state compression. The update mechanism plays an important role for stability and should not be compressed.
Compared to the Fastfood transform \cite{yang2015deep}, our method is better at compressing entire cells. It achieves higher accuracy with the same number of weights both in adding problem and copy memory problem. For example, in adding problem, our method achieves a higher accuracy of 91.64\% (\emph{vs.} 85.99\%) with the same number of weights, compared to the Fastfood transform \cite{yang2015deep}.

\subsubsection{Sequential-MNIST}
The sequential MNIST benchmark dataset has previously been described in Section~\ref{seq:cnn_mnist}. A gray-scale image with a size of 28$\times$28 is interpreted as a sequence of 784 pixels. 
The entire sequence is an input to the GRU, which will generate a classification score.
We select a GRU with a hidden size of 512 as our baseline and an RMSProp optimizer with a learning rate of 0.001.
\begin{table}[t]
\caption{RNN compression results on the sequential MNIST and Penn-Treebank benchmarks. On the sequential MNIST benchmark (a), the pattern here reflects what we saw on the adding and copy-memory benchmarks. Touching the update gate has a negative impact. All other equations can be compressed. our method (WaveGRU-64) achieves a comparable performance, compared to \cite{tjandra2017compressing}. In (b) we show results for the best performing architectures on the Penn-Treebank data set, we compare to a TCN as proposed in \cite{BaiTCN2018}. We can compress the GRU cells state and reset equations without a significant drop in performance.}
\begin{subfigure}[t]{0.54\textwidth}
\centering
\caption{Sequential MNIST}
\begin{tabular}{ c c c c c}
            & reduced & loss   & accuracy  & weights  \\  \toprule
GRU         & -       & 6.49e-2  & 100\%    & 795K \\
Wave-GRU    & $\mathbf{z_t}$   & 8.98e-2 & 98\% & 534K \\
Wave-GRU    & $\mathbf{g_r}$   & 6.06e-2 & 100\% & 534K \\
Wave-GRU    & $\mathbf{g_z}$   & 1.82 & 26\% & 534K \\
Wave-GRU    & $\mathbf{g_r}$, $\mathbf{g_z}$ & 1.33 &  46\%  & 274K \\ 
Wave-GRU    & $\mathbf{z_t}$, $\mathbf{g_r}$ & 9.48e-2 & 98\% & 274K\\
Wave-GRU    & $\mathbf{z_t}$, $\mathbf{g_z}$   & 1.60 & 34\%  & 274K\\
Wave-GRU    & $\mathbf{z_t}$, $\mathbf{g_z}$, $\mathbf{g_r}$   & 1.52 & 36\%  & 13K\\ \midrule
WaveGRU-64  &  $\mathbf{z_t}$, $\mathbf{g_r}$  & 0.127  & 96.4\% &  4.9K \\ 
TT-GRU & -     & -  & 98.3\%    & 5.1K \\
\bottomrule
\end{tabular}
\label{tab:seqMnist}
\end{subfigure}
\begin{subfigure}[t]{0.4\textwidth}
\centering
\caption{Penn-Treebank}
\begin{tabular}{c c c c c}
            & reduced & loss   & bpc  & weights  \\  \toprule
TCN         & -       & 0.916  & 1.322 & 2,221K \\
GRU         & -       & 0.925  & 1.335 & 972K \\
Wave-GRU    & $\mathbf{z_t}$   & 0.97  & 1.399 & 711K \\
Wave-GRU    & $\mathbf{g_r}$   & 0.925 & 1.335 & 711K \\
Wave-GRU    & $\mathbf{z_t}$, $\mathbf{g_r}$ & 0.969 & 1.398 & 450K \\
\bottomrule
\end{tabular}
\label{tab:PennCompression}
\end{subfigure}
\end{table}
Results of our method are shown in Table~\ref{tab:seqMnist}.
Similar to the results of Table \ref{tab:adding_memory}, we also observe that having efficient representations for the state and reset gate matrices work well, but compressing the update gate adversely impacts the results. 
Compared to only state compression, the combination of state and reset compression achieves a higher compression rate of 2.9$\times$ (\emph{vs.} 1.5$\times$), without the accuracy drop. 
We also compare to the tensor train approach used in \cite{tjandra2017compressing}. We apply the efficient wavelet layer only on the reset and state weight matrices and reduce the cell size to 64. Our approach does reasonably well with fewer parameters.

\subsubsection{Penn Treebank Character Modelling}
We verify our approach on the Penn-Treebank (PTB) character modelling benchmark \cite{marcus1993building}. 
We split the dataset into training, validation and test sequences, which contains 5,059K training characters, 396K validation characters and 446K testing characters. 
Given an input sequence of 320 characters, the model should predict the next character. We work with a GRU of size 512 trained using an Adam optimizer with an initial learning rate of 0.005. Training is done using a cross entropy loss in addition to the wavelet loss, and results are reported using bits per character (bpc), where lower bpc is better.
In Table~\ref{tab:PennCompression}, we show results for a temporal convolutional network (TCN) \cite{BaiTCN2018}, a vanilla GRU cell as well as state, reset and state reset compression, which we found to be successfully earlier. We confirm that our wavelet-based compression method can be used to compress reset gate and cell state without significant performance loss.

\section{Conclusion}
We presented a novel wavelet based efficient linear layer which demonstrates competitive performance within convolutional and recurrent network structures. On the MNIST digit recognition benchmark, we show state of the art compression results as well as convergence from randomly initialized filters.
We explore RNN compression and observe comparable performance on the sequential MNIST task. In a gated recurrent unit we can compress the reset and state equations without a significant impact on performance. The update gate equation was hard to compress, in particular in combination with the state equation. Joint update gate and reset gate equation compression generally worked better than update and state compression. We conclude that the update mechanism plays the most important role within a GRU-cell, followed by the state equation and finally the reset gate. Results indicate that selective compression can significantly reduce cell parameters while retaining good performance.
Product filters are only one way of wavelet design, alternative methods include lifting or spectral factorization approaches. We look forward to exploring some of these in the future.  Efficient implementation of the FWT on GPUs is no simple matter. We hope this paper will spark future work on highly optimized implementations. 

\smallskip
\footnotesize{\noindent\textbf{Acknowledgements:} Research was supported by the Deutsche Forschungsgemeinschaft (DFG, German Research Foundation) project YA 447/2-1 (FOR 2535 Anticipating Human Behavior) and by the National Research Foundation Singapore under its NRF Fellowship Programme [NRF-NRFFAI1-2019-0001].}

{
\bibliographystyle{unsrt}
\bibliography{bib.bib}
}
\end{document}